\documentclass[pdflatex,sn-mathphys-num]{sn-jnl}

\usepackage{graphicx}
\usepackage{multirow}
\usepackage{amsmath,amssymb,amsfonts}
\usepackage{xcolor}
\usepackage{textcomp}
\usepackage{booktabs}
\usepackage{subcaption}
\usepackage{makecell}
\usepackage{pgfplots}
\pgfplotsset{compat=1.18}

\definecolor{apricot}{RGB}{0,204,0}
\graphicspath{{images/}}

\begin{document}

\title[PriorNet: Prior-Guided Engagement Estimation from Face Video]{PriorNet: Prior-Guided Engagement Estimation from Face Video}

\author*{\fnm{Alexander} \sur{Vedernikov}}\email{alexander.vedernikov.edu@gmail.com}
\affil{\orgaddress{\city{Helsinki}, \country{Finland}}}

\abstract{Engagement estimation from face video remains challenging because facial evidence is often incomplete, labeled data are limited, and engagement annotations are subjective. We present PriorNet, a prior-guided framework that injects task-relevant priors at three stages of the pipeline: preprocessing, model adaptation, and objective design. PriorNet converts face-detection failures into explicit zero-frame placeholders so that missing-face events remain represented in the input sequence, adapts a frozen Self-supervised Video Facial Affect Perceiver (SVFAP) backbone through a Prior-guided Low-Rank Adaptation module (Prior-LoRA) for parameter-efficient specialization, and trains with a Dirichlet-evidential, uncertainty-weighted objective under hard-label supervision. We evaluate PriorNet on EngageNet, DAiSEE, DREAMS, and PAFE using each dataset's native evaluation protocol. Across these benchmarks, PriorNet improves over the strongest listed prior reference within each dataset's evaluation framing, while component ablations on EngageNet and DAiSEE indicate that the gains arise from complementary contributions of preprocessing, adaptation, and objective-level priors. These results support explicit prior injection as a useful design principle for face-video engagement estimation under the benchmark conditions studied in this work.}

\keywords{Affective Computing, Engagement, Facial Analysis, Video Understanding}

\maketitle


\section{Introduction}
\label{sec:intro}

Estimating engagement from face video is important for machine-vision applications such as online learning analytics, virtual meetings, and human-robot interaction, where systems must infer user involvement from subtle visual behavior rather than explicit feedback \cite{karimah2022automatic,singh2023do,Singh2024dreams}. Yet the task remains difficult under practical conditions. In realistic video streams, facial evidence is often incomplete: users look away from the screen, rotate their heads, move partially out of frame, or are observed under low resolution, occlusion, and unstable illumination \cite{kumar2025comput}. At the same time, engagement labels are coarse and subjective, and available benchmarks remain modest in size relative to modern video backbones. As a result, engagement estimation from face video is not only a representation-learning problem, but also a robustness problem.

Recent work has improved engagement prediction through stronger temporal encoders, graph-based reasoning, and transformer-style video models \cite{abedi2021improving,singh2023do,abedi2024eng,ai2025cavt}. These advances have clearly moved the field beyond handcrafted descriptors and shallow classifiers. However, three limitations remain. First, missing-face events are usually treated as preprocessing failures or nuisance corruption, even though in engagement analysis they may reflect gaze aversion, head turning, or other behaviorally meaningful states. Second, model adaptation is typically framed as full fine-tuning or feature extraction with task-specific heads, despite the small-data regime of most engagement benchmarks. Third, supervision is still largely based on standard or mildly structured classification objectives, while the ambiguity inherent in engagement labels is only weakly modeled. Consequently, current methods improve benchmark performance, but they do not systematically address the conditions under which face-video engagement estimation becomes unreliable.

In this work, we address these issues through a prior-guided engineering perspective. Our premise is that robustness improves when task-relevant priors are injected explicitly at complementary stages of the pipeline rather than left to emerge implicitly from a larger model. Based on this view, we propose PriorNet, a framework that combines three design choices aligned with the main failure modes of the task: (i) zero-frame placeholders that encode face-detection failures as explicit input evidence, (ii) Prior-LoRA, a prior-guided low-rank adaptation module, that specializes a frozen Self-supervised Video Facial Affect Perceiver (SVFAP) backbone with a small trainable budget \cite{hu2021lora}, and (iii) a Dirichlet-evidential and uncertainty-weighted objective that better handles ambiguous or subjective supervision under hard-label training. The goal of PriorNet is not to replace strong pretrained video representations, but to bias them toward the regimes in which engagement estimation is most brittle.

We evaluate PriorNet on four heterogeneous engagement benchmarks (EngageNet, DAiSEE, DREAMS, and PAFE) using each dataset's native evaluation protocol. Across these settings, PriorNet improves over the strongest comparable prior results, while component ablations on EngageNet and DAiSEE show that the gains are attributable to the proposed preprocessing, adaptation, and objective priors rather than to a single isolated design choice. Taken together, these results support our central claim: for engagement estimation from face video, robustness is better achieved by explicit prior injection than by architecture scaling alone. The contributions of this work are: 

\begin{enumerate}
    \item \textbf{Prior-guided robustness framework.} We formulate engagement estimation from face video as a prior-injection problem and present PriorNet, a unified framework that injects task-relevant priors at the preprocessing, adaptation, and objective levels.

    \item \textbf{Explicit modeling of missing-face events.} We convert face-detection failures into zero-frame placeholders, enabling the model to treat missing-face events as informative behavioral evidence rather than discarded corruption.

    \item \textbf{Parameter-efficient specialization for engagement video.} We introduce Prior-LoRA within a frozen SVFAP backbone to adapt spatio-temporal representations with less than 1\% trainable parameters, targeting robust learning under limited labeled data.

    \item \textbf{Uncertainty-aware learning for ambiguous labels.} We optimize PriorNet with a Dirichlet-evidential and uncertainty-weighted objective that is designed to improve robustness under subjective engagement supervision.

    \item \textbf{Cross-dataset validation and module-level analysis.} We show consistent improvements across EngageNet, DAiSEE, DREAMS, and PAFE, and provide component ablations on EngageNet and DAiSEE, together with a placeholder-focused diagnostic analysis on EngageNet.
\end{enumerate}


\section{Related Work}
\label{sec:2_related_work}

\subsection{From engineered cues to end-to-end engagement video models}
Early automatic engagement-estimation systems relied on engineered behavioral cues (facial expressions, gaze, posture, or physiological surrogates) combined with shallow classifiers. These works established that engagement is at least partly observable from human behavior, but they were sensitive to pose, illumination, occlusion, and sensor availability \cite{Whitehill201486,nezami2019auto,monkaresi2017auto}. More recent methods reformulated the problem as spatio-temporal video understanding, replacing handcrafted descriptors with CNN--RNN, TCN, graph, or transformer-based models that learn temporal structure directly from clips or derived behavioral streams \cite{abedi2021improving,singh2023do,abedi2024eng,ai2025cavt}. In this sense, the field has clearly progressed from static facial analysis to richer sequence modeling.

However, most of this progression has focused on stronger temporal encoders, larger backbones, or more expressive fusion strategies, while paying less attention to engagement-specific failure modes. In particular, current pipelines still tend to treat incomplete facial evidence as nuisance corruption and to optimize against ambiguous labels with standard objectives. \textbf{Thus, the central unresolved problem is no longer whether engagement can be modeled from video, but how to make video-based engagement estimation robust to missing facial evidence, limited task-specific data, and subjective annotations.}

\subsection{Preprocessing priors: stabilizing visible cues versus modeling missing-face events}
A natural way to improve robustness is to inject priors before representation learning. In engagement and adjacent affective-computing pipelines, preprocessing is commonly used to normalize visible cues through face cropping, landmark extraction, head-pose estimation, gaze tracking, or multimodal alignment. Toolkits such as OpenFace 2.0 have made this style of preprocessing practical by providing facial landmarks, head pose, action units, and gaze estimates from unconstrained video \cite{baltrusaitis2018openface2}. Other engagement-oriented systems further combine video with audio, interaction logs, or physiological signals, but such pipelines often require more instrumentation, more engineering, or stronger assumptions about input quality \cite{aslan2019investigating}.

A smaller body of adjacent work has explored explicit handling of missing observations via masks, zeroed inputs, or placeholder signals \cite{psaltis2017multimodal,khenkar2022engagement}. Yet these approaches were not developed as a direct solution to face-video engagement estimation; more importantly, they typically frame missingness as nuisance corruption or as a generic masking problem rather than as semantically informative behavior. For engagement analysis, this distinction matters: a failed face detection can reflect gaze aversion, head turning, partial leaving of the camera view, or other visually weak but behaviorally meaningful states. \textbf{What remains missing is a preprocessing prior that does not merely compensate for missing facial evidence, but explicitly encodes missing-face events as part of the engagement signal itself.}

\subsection{Adaptation priors: task-specific engagement architectures versus parameter-efficient specialization}
A second line of work injects priors inside the model architecture. Within engagement estimation, this has mainly taken the form of task-specific temporal designs: CNN-TCN hybrids, ordinal graph models, and hybrid convolution--transformer networks that bias the representation toward temporal dynamics or the ordinal structure of engagement labels \cite{abedi2021improving,abedi2023affect,abedi2024eng,Vedernikov_2024_CVPR}. These models show that architectural bias matters, especially when engagement cues are subtle and short-lived.

At the same time, the broader vision literature has shown that large pretrained transformers need not be adapted by updating the full backbone. AdaptFormer demonstrated that lightweight adaptation modules can transfer pretrained vision transformers to image and video tasks efficiently, while more recent low-rank fine-tuning work has refined how task-specific updates can be introduced without drifting too far from pretrained representations \cite{chen2022adaptformer,dong2024lowrank}. These ideas are especially relevant for engagement benchmarks, where labeled data are relatively limited and full fine-tuning can easily overfit. Yet parameter-efficient fine-tuning has remained largely underexplored in engagement video, where adaptation is still usually framed either as full model tuning or as adding task-specific heads on top of frozen features. \textbf{The resulting gap is a lack of engagement-specific parameter-efficient adaptation that can specialize a strong pretrained video backbone to subtle facial dynamics without sacrificing the robustness of its pretrained priors.}

\subsection{Objective priors: structured engagement supervision without explicit ambiguity modeling}
The final unresolved axis concerns the learning objective. Engagement labels are often coarse, subjective, and protocol-dependent: they may come from external annotators, self-reports, ordinal categories, or crowd aggregation. Prior engagement work has partially acknowledged this by using ordinal formulations, temporal smoothness assumptions, or multiple-instance learning to better reflect the structure of engagement annotations \cite{kaur2018predict,abedi2023affect}. These are important steps beyond plain frame-wise classification.

Nevertheless, they still stop short of treating label ambiguity as first-class signal. In broader uncertainty-aware learning, evidential deep learning models class predictions as Dirichlet evidence rather than point probabilities \cite{sensoy2018evidential}. More recent work extends this perspective to composite or vague labels via Hyper-Evidential Neural Networks (HENN) \cite{li2024hyper}, and calibration-oriented learning has shown that uncertainty-weighted gradients can improve the reliability of predictive confidence by explicitly emphasizing uncertain samples \cite{lin2025uncertainty}. However, these developments have not been systematically incorporated into face-video engagement estimation, where ambiguity is intrinsic to the target rather than incidental annotation noise. \textbf{Accordingly, the field still lacks an uncertainty-aware engagement objective that is explicitly designed for ambiguous labels and integrated with a modern video-adaptation pipeline.}

\vspace{0.25cm}
Taken together, the literature points to three complementary but still disconnected needs: explicit treatment of missing-face events, parameter-efficient specialization of pretrained video backbones, and uncertainty-aware learning for ambiguous engagement labels. Existing methods address these axes only in isolation, or outside the specific setting of face-video engagement estimation. \textbf{PriorNet is motivated by exactly this gap: it unifies preprocessing priors, adaptation priors, and objective priors within a single robustness-oriented framework.}


\section{Method}\label{sec:method}

\subsection{PriorNet Overview}
\label{sec:PriorNet}

\begin{figure*}[h]
  \centering
   \includegraphics[width=1.\linewidth]{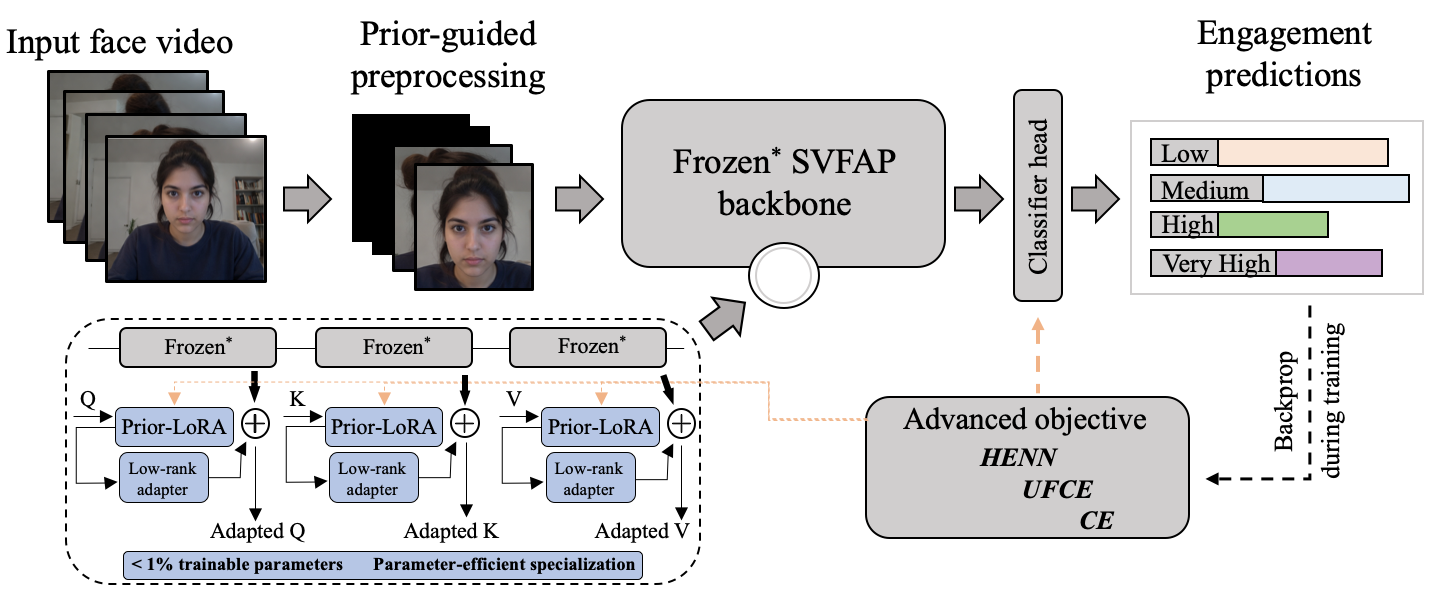}
   \caption{Overview of PriorNet. Input face video is converted into a fixed-length clip with zero-frame placeholders for missing-face events. A frozen SVFAP backbone with Prior-LoRA adapters in selected Q/K/V projections produces dataset-specific predictions, trained with the proposed uncertainty-aware objective.} 
   \label{fig:method_main}
\end{figure*}

PriorNet injects task-relevant priors at three stages of the engagement-estimation pipeline: preprocessing, model adaptation, and objective design (see Fig.~\ref{fig:method_main}). Given an input face video, we first construct a fixed-length facial clip using the preprocessing procedure described in Sec.~\ref{sec:preprocessing}. Frames for which face detection fails are replaced with zero-frame placeholders so that missing-face events remain explicitly represented rather than being discarded.

The resulting clip is then processed by SVFAP, a self-supervised facial-video backbone pretrained by masked facial video autoencoding with a temporal pyramid and spatial bottleneck transformer architecture \cite{sun2024svfap}. We choose SVFAP as the base encoder because it is pretrained specifically on facial-video affect dynamics, making it a more suitable foundation for engagement estimation than generic video backbones while still requiring task-specific adaptation for robust engagement prediction. To adapt this backbone to engagement estimation under limited labeled data, we keep the pretrained SVFAP weights frozen and insert Prior-LoRA modules into its self-attention projections. This design enables task-specific specialization with a small trainable parameter budget while preserving the backbone's pretrained affective representations.

Finally, engagement predictions are optimized with the uncertainty-aware objective defined in Sec.~\ref{sec:uncertainty}. This objective combines evidential supervision with uncertainty-weighted reweighting to better accommodate ambiguous and subjective engagement labels. In this way, PriorNet couples explicit modeling of missing-face events, parameter-efficient backbone adaptation, and uncertainty-aware optimization within a single face-video engagement framework. Because the three priors are introduced at the preprocessing, adaptation, and objective levels, the proposed design is not tied to a single classification head; in principle, the same formulation can be instantiated with other facial-video backbones, although this work evaluates it only with SVFAP. The following subsections detail each component in turn.

\subsection{Engagement-Specific Preprocessing}
\label{sec:preprocessing}

We first sample \(N=16\) frames uniformly across the temporal extent of the video in the spirit of the sparse sampling strategy used in Temporal Segment Networks \cite{wang2016temporal}. The sampled frame indices are
\begin{equation}
t_i = \left\lfloor \frac{(i-1)(T-1)}{N-1} \right\rfloor + 1,
\qquad i=1,\dots,N,
\label{eq:frame_sampling}
\end{equation}
where \(T\) is the total number of frames in the input video.

\begin{figure}[h]
  \centering
   \includegraphics[width=1\linewidth]{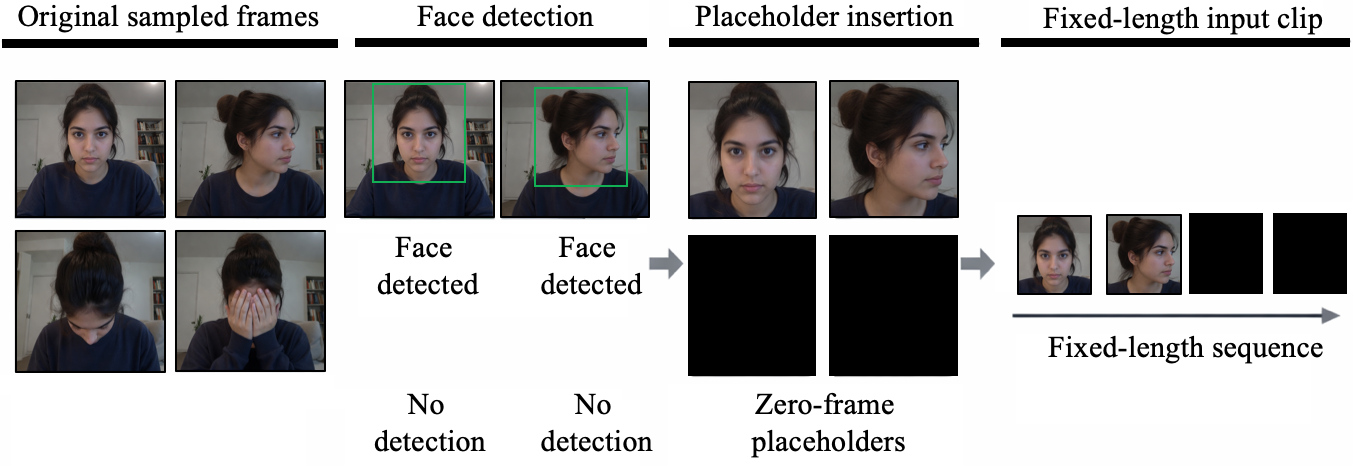} 
   \caption{Prior-guided preprocessing in PriorNet. Sampled frames are processed by a face detector; detected faces are cropped, while frames with failed detection are replaced by zero-frame placeholders. This yields a fixed-length clip in which missing-face events are explicitly encoded for downstream engagement estimation.} 
   \label{fig:preprocessing} 
\end{figure}

For each sampled frame \(x_{t_i}\), we apply an OpenCV SSD-based face detector with a ResNet-10 backbone and confidence threshold 0.5 \cite{liu2016ssd}. We use this detector as a lightweight and reproducible preprocessing component for frame-wise face localization. If a face is detected, the detector returns a bounding box \(b_{t_i}\); otherwise \(b_{t_i}=\varnothing\). The corresponding preprocessed frame is then defined as
\begin{equation}
y_i =
\begin{cases}
\mathcal{R}\bigl(\mathrm{Crop}(x_{t_i}, b_{t_i})\bigr), & b_{t_i} \neq \varnothing,\\[4pt]
\mathbf{0}_{224 \times 224 \times 3}, & b_{t_i} = \varnothing,
\end{cases}
\label{eq:preproc_output}
\end{equation}
where \(\mathrm{Crop}(x,b)\) extracts the image region inside bounding box \(b\), and \(\mathcal{R}(\cdot)\) resizes the crop to \(224 \times 224\) by bilinear interpolation to match the SVFAP input resolution.

This design preserves a fixed clip length regardless of whether facial evidence is available in every sampled frame. When detection succeeds, the model receives a face-centered crop; when detection fails, the corresponding time step is retained as an all-zero placeholder rather than being removed. As a result, missing-face events remain explicitly observable to the downstream model. We do not assume that such events correspond deterministically to disengagement; instead, the placeholder marks missing facial visibility, which may co-occur with engagement-relevant behaviors such as gaze diversion, strong head rotation, partial out-of-frame motion, or challenging visual conditions. The model can then learn whether these events are informative for engagement estimation in context.

\subsection{Prior-LoRA for Efficient Fine-Tuning}
\label{sec:peft}

\begin{figure}[h]
  \centering
   \includegraphics[width=1\linewidth]{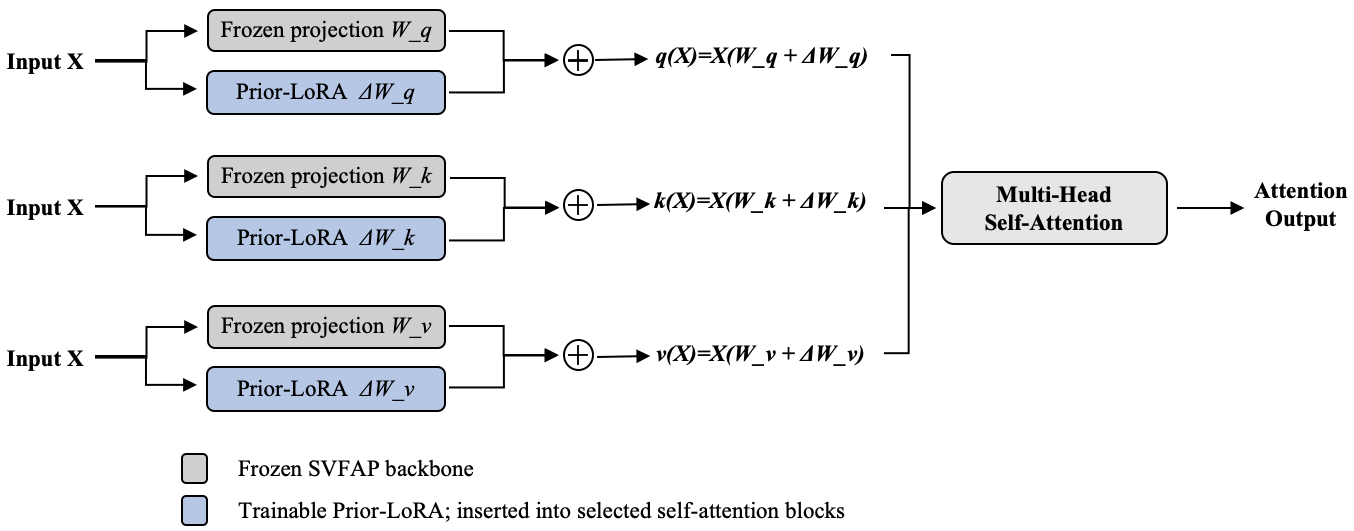} 
   \caption{Proposed Prior-LoRA adaptation inside the transformer attention block. Trainable low-rank adapters are added to the query, key, and value projections while the original backbone remains frozen, enabling parameter-efficient specialization for engagement estimation.} 
   \label{fig:lora}
\end{figure}

To adapt the pretrained SVFAP backbone to engagement estimation under limited labeled data, we employ Low-Rank Adaptation (LoRA) \cite{hu2021lora} as a parameter-efficient alternative to full fine-tuning. The key idea is to keep the backbone weights frozen and introduce trainable low-rank residual updates only inside selected self-attention projections. We refer to this mechanism as a prior-guided low-rank adaptation module (Prior-LoRA) because the adapters are inserted into selected self-attention projections of the SVFAP encoder, enabling task-specific refinement of facial-video representations while preserving the backbone's pretrained affective priors.

Let \(X \in \mathbb{R}^{L \times d}\) denote the token matrix entering a self-attention layer, where \(L\) is the number of tokens and \(d\) is the hidden dimension. The frozen query, key, and value projections are parameterized by
\[
W_q,\;W_k,\;W_v \in \mathbb{R}^{d \times d}.
\]
For each selected attention layer, Prior-LoRA introduces trainable low-rank residual updates
\begin{equation}
\Delta W_q = \frac{\alpha}{r} A_q B_q,\qquad
\Delta W_k = \frac{\alpha}{r} A_k B_k,\qquad
\Delta W_v = \frac{\alpha}{r} A_v B_v,
\label{eq:priorlora_delta}
\end{equation}
where
\[
A_* \in \mathbb{R}^{d \times r}, \qquad
B_* \in \mathbb{R}^{r \times d}, \qquad
r \ll d.
\]
The adapted projections are then
\begin{equation}
\begin{aligned}
q(X) &= X\bigl(W_q + \Delta W_q\bigr),\\
k(X) &= X\bigl(W_k + \Delta W_k\bigr),\\
v(X) &= X\bigl(W_v + \Delta W_v\bigr).
\end{aligned}
\label{eq:priorlora}
\end{equation}

In the default configuration, Prior-LoRA is applied to every other self-attention block of the SVFAP encoder with rank \(r=16\). The low-rank residual branch is initialized to produce zero output at the start of training, so the model initially behaves identically to the pretrained backbone. During fine-tuning, only the LoRA parameters and the final classifier head are optimized, while the original SVFAP backbone weights remain frozen. This reduces the number of trainable parameters to below \(1\%\) of the backbone.

From a modeling perspective, Prior-LoRA does not replace the pretrained spatio-temporal representations learned by SVFAP; instead, it adds a lightweight task-specific correction inside the attention mechanism. This is particularly desirable for engagement estimation, where benchmark datasets are modest in size and full fine-tuning is more prone to overfitting or to overwriting pretrained affective structure.

\subsection{Loss Computation via Dirichlet-Evidential and Uncertainty-Weighted Objectives}
\label{sec:uncertainty}

To improve robustness under ambiguous and noisy engagement labels, we employ a classification objective that combines a Dirichlet-evidential term with an uncertainty-weighted focal-like term. This objective operates on hard class labels rather than annotator-vote distributions. Thus, uncertainty is modeled through evidence derived from the network outputs, while supervision remains standard class-index supervision.

\vspace{0.1cm}
\noindent \textbf{Dirichlet-Evidential Term.}
We adopt this evidential formulation because Dirichlet-based classification provides a principled way to represent both class preference and predictive uncertainty within a single deterministic model, and recent hyper-evidential extensions further motivate its use when class boundaries are ambiguous \cite{sensoy2018evidential,li2024hyper}. Let \(\mathbf{z}_i \in \mathbb{R}^{C}\) denote the pre-softmax logits for sample \(i\). We first compute non-negative evidence:
\begin{equation}
\mathbf{e}_i = \operatorname{clip}\!\left(\operatorname{softplus}(\mathbf{z}_i),\,0,\,5\right),
\qquad
\boldsymbol{\alpha}_i = \mathbf{e}_i + \mathbf{1},
\qquad
\alpha_{0,i} = \sum_{k=1}^{C}\alpha_{i,k},
\label{eq:henn_evidence}
\end{equation}
where \(C\) is the number of classes. Following evidential classification, the data term for the ground-truth class \(y_i\) is written in closed form as:
\begin{equation}
\mathcal{L}^{(i)}_{\mathrm{data}}
=
\psi(\alpha_{0,i})-\psi(\alpha_{i,y_i}),
\label{eq:henn_data}
\end{equation}
where \(\psi(\cdot)\) is the digamma function. This term is complemented by a Dirichlet regularizer:
\begin{equation}
\mathcal{L}^{(i)}_{\mathrm{KL}}
=
\mathrm{KL}\!\left[\mathrm{Dir}(\boldsymbol{\alpha}_i)\,\|\,\mathrm{Dir}(\mathbf{1})\right].
\label{eq:henn_kl}
\end{equation}
The resulting evidential component is:
\begin{equation}
\mathcal{L}^{(i)}_{\mathrm{HENN}}
=
\mathcal{L}^{(i)}_{\mathrm{data}}
+
\lambda_{\mathrm{KL}}\,\mathcal{L}^{(i)}_{\mathrm{KL}}.
\label{eq:henn_loss}
\end{equation}
While the evidential term models predictive uncertainty, we further introduce an uncertainty-weighted focal-like term to place greater emphasis on uncertain and difficult samples during optimization \cite{lin2025uncertainty}.

\vspace{0.1cm}
\noindent \textbf{Uncertainty-Weighted Focal-Like Term.}
We introduce this term to reduce overconfident learning on ambiguous engagement samples by assigning greater optimization weight to predictions that remain uncertain under the evidential model. In this sense, the formulation is related to focal-style reweighting, but uses uncertainty-dependent modulation rather than a fixed exponent \cite{lin2017focal,lin2025uncertainty}. We first compute class probabilities:
\begin{equation}
\mathbf{p}_i = \operatorname{softmax}(\mathbf{z}_i),
\label{eq:ufce_softmax}
\end{equation}
and define a sample-wise uncertainty weight from the total evidence:
\begin{equation}
u_i = \frac{1}{1 + \sum_{k=1}^{C} e_{i,k}}.
\label{eq:ufce_uncertainty}
\end{equation}
The uncertainty-weighted focal-like term is then:
\begin{equation}
\mathcal{L}^{(i)}_{\mathrm{UFCE}}
=
-
w_{\mathrm{ufce}}\,
u_i\,
\bigl(1-p_{i,y_i}\bigr)^{u_i}
\log\bigl(p_{i,y_i}+\varepsilon\bigr),
\label{eq:ufce_loss}
\end{equation}
where \(\varepsilon\) is a small numerical constant. Unlike standard focal loss, the exponent is not a fixed hyperparameter; instead, it is modulated by the sample uncertainty \(u_i\).

\vspace{0.1cm}
\noindent \textbf{Combined Objective.}
Our final advanced objective combines the evidential term, the uncertainty-weighted focal-like term, and an auxiliary hard-label cross-entropy term for optimization stability. Over a mini-batch of size \(B\), it is defined as
\begin{equation}
\mathcal{L}_{\mathrm{full}}
=
\frac{1}{B}\sum_{i=1}^{B}
\left(
\mathcal{L}^{(i)}_{\mathrm{HENN}}
+
\mathcal{L}^{(i)}_{\mathrm{UFCE}}
+
w_{\mathrm{ce}}\,\mathcal{L}^{(i)}_{\mathrm{CE}}
\right),
\label{eq:combined_loss_full}
\end{equation}
where
\begin{equation}
\mathcal{L}^{(i)}_{\mathrm{CE}} = -\log\bigl(p_{i,y_i}+\varepsilon\bigr).
\label{eq:ce_loss}
\end{equation}

The auxiliary cross-entropy term preserves a direct discriminative training signal under hard-label supervision, while the evidential and uncertainty-weighted terms model predictive uncertainty and place greater emphasis on ambiguous and difficult samples.


\section{Experimental Setup}
\label{sec:experimental_setup}

\subsection{Datasets}
\label{sec:datasets}

We evaluate PriorNet on four engagement-related facial-video benchmarks that differ substantially in label space, split design, and published evaluation protocol: EngageNet \cite{singh2023do,dhall2023emotiw}, DAiSEE \cite{gupta2016daisee}, DREAMS (Diverse Reactions of Engagement and Attention Mind States) \cite{Singh2024dreams}, and PAFE (Predicting Attention with Facial Expression) \cite{Taeckyung2022PAFE}. We preserve each dataset's native task formulation rather than forcing a single unified protocol across all benchmarks.

\vspace{.04in}\noindent
\textbf{EngageNet} is a large-scale in-the-wild engagement benchmark containing 11{,}311 facial-video clips from 127 users aged $18$ to $37$, annotated into four engagement levels: \emph{Highly Engaged}, \emph{Engaged}, \emph{Barely Engaged}, and \emph{Not Engaged}. The official split is subject-independent and consists of 7{,}983 training clips ($90$ users), 1{,}071 validation clips ($11$ users), and 2{,}257 test clips ($26$ users). Each clip was independently annotated by three experts.

\vspace{.04in}\noindent
\textbf{DAiSEE} contains 9{,}068 video clips from 112 participants and provides labels for boredom, confusion, engagement, and frustration at four ordinal levels. In this work, we consider the engagement label only, following the standard engagement-classification usage of the dataset.

\vspace{.04in}\noindent
\textbf{DREAMS} contains facial videos of 32 participants recorded in naturalistic conditions while they watch diverse stimuli designed to evoke different reactions. In total, $832$ video segments of $40$ seconds each were recorded, with $781$ segments successfully used for analysis. Every 40 seconds, participants annotated their engagement levels via self-assessment prompts. The benchmark studies both engagement and attention, and the original paper reports separate single-task, transfer-learning, and multi-task settings.

\vspace{.04in}\noindent
\textbf{PAFE} is an in-the-wild facial-video benchmark for attentional-state prediction during online lectures. Each sample was recorded at $640$p resolution and $30$fps, with probes conducted every $40$ seconds to label attentional states as `Focused,' `Not-Focused,' or `Skip.'  After data cleaning, the released benchmark contains approximately 15 hours of recordings from 15 participants and 1{,}100 attention probes, with labels derived from probe-caught self-reports.

\subsection{Evaluation Protocol}
\label{sec:evaluation_protocol}
Because these four benchmarks differ in split structure, target definition, and published reporting conventions, we do not aggregate them under a single artificial protocol. Instead, we follow each dataset's native evaluation setting and restrict direct comparison to prior results reported under matching split, task, and metric conditions. Methods evaluated under substantially different protocols are discussed where relevant but are not treated as directly rank-comparable in the main benchmark tables.

\vspace{.04in}\noindent
\textbf{EngageNet.} We follow the official subject-independent split introduced with the dataset. However, because public benchmark comparisons are commonly reported on the validation partition rather than on hidden test labels, we use the validation split as the public comparison split for EngageNet. Accordingly, EngageNet numbers in this manuscript should be interpreted as validation-set benchmark results for direct comparison with prior public reports, rather than as hidden-test performance. Classification accuracy is used as the primary comparison metric.

\vspace{.04in}\noindent
\textbf{DAiSEE.} We use the official subject-exclusive train/validation/test partition provided with the dataset. Following common fixed-split evaluation practice, the training and validation subsets are merged for model fitting, and final performance is reported on the held-out test set. We use classification accuracy as the primary comparison metric for the engagement label.

\vspace{.04in}\noindent
\textbf{DREAMS.} We follow the subject-independent classification setting introduced in the original dataset paper. Because DREAMS reports single-task, transfer-learning, and multi-task settings separately, comparisons are restricted to the matching setting reported in the corresponding benchmark table and are not pooled across incompatible regimes. Weighted F1 is used as the primary comparison metric for direct benchmarking.

\vspace{.04in}\noindent
\textbf{PAFE.} We follow the participant-disjoint stratified five-fold cross-validation protocol used in the original benchmark. The model is trained and evaluated independently in each fold, and mean performance across folds is reported. As in the original work, weighted F1 is used as the primary comparison metric.

\subsection{Training Protocol}
\label{sec:training_protocol}
All PriorNet experiments use the preprocessing pipeline of Sec.~\ref{sec:preprocessing}, the SVFAP backbone with Prior-LoRA adaptation from Sec.~\ref{sec:peft}, and the objective defined in Sec.~\ref{sec:uncertainty}, unless a simpler loss is intentionally used for baseline variants in the ablation study. For the ablation baseline, standard cross-entropy is used instead of the advanced objective.

The optimization settings are selected separately for each dataset to account for differences in dataset size, split design, label structure, and evaluation protocol. Within a given dataset, the same training recipe is kept fixed across the compared PriorNet variants unless a specific ablation explicitly changes one component of the method. Zero-frame placeholders and Prior-LoRA remain active during both training and evaluation. For cross-validation settings such as PAFE, the training procedure is repeated independently in each fold and the final metric is averaged across folds.

To ensure fair benchmarking, preprocessing, architecture, and optimization choices are kept fixed within each experiment block. Any comparison to prior work is therefore interpreted under the published protocol of the corresponding dataset rather than as a cross-dataset pooled score. Component ablations are reported on EngageNet and DAiSEE under their respective native benchmark settings, while the placeholder-rate diagnostic analysis is reported on EngageNet only.


\section{Results}
\label{sec:results}

The dataset-specific protocols used for evaluation are defined in Sec.~\ref{sec:experimental_setup}. In this section, we report results under those native benchmark settings and restrict direct comparison to prior work reported under matching split, task, and metric conditions. When published settings are not fully uniform, the corresponding table is intended as protocol-aware benchmark context rather than as a strictly homogeneous leaderboard. Taken together, these benchmark results establish the empirical scope of the improvement, while Sec.~\ref{sec:ablation_analysis} examines which components of PriorNet are responsible for that gain.

\subsection{EngageNet}

Table~\ref{tab:method_principle_accuracy} reports EngageNet results under the public validation-split benchmark setting described in Sec.~\ref{sec:evaluation_protocol}. Only methods reported on that public validation regime are treated as directly comparable in the main ranking; studies using cross-validation, task reformulation, or other non-matching protocols are excluded from the table. Recent EngageNet-related studies using non-matching evaluation settings, such as cross-validation or task reformulation, are not treated as directly comparable in this benchmark context \cite{jiang2025prompting,computers14030109}.

\begin{table}[h]
  \centering
  \caption{Comparative analysis of representative methods and PriorNet accuracy on EngageNet.}
  \label{tab:method_principle_accuracy}
  {\fontsize{8.1}{9.5}\selectfont
  \setlength\tabcolsep{6pt}
  \begin{tabular}{@{}l l c@{}}
  \toprule
    \textbf{Method} & \textbf{Principle} & \textbf{Accuracy [\%]} \\
    \midrule
    ResNet + TCN \cite{abedi2021improving} & End-to-end & 54.72 \\
    EfficientNet + LSTM \cite{selim2022students} & End-to-end & 57.57 \\
    EfficientNet + Bi-LSTM \cite{selim2022students} & End-to-end & 58.94 \\

    CNN + LSTM \cite{singh2023do} & Sequential features & 67.51 \\
    LSTM \cite{singh2023do} & Sequential features & 67.69 \\
    TCN \cite{singh2023do} & Temporal features & 67.79 \\
    VLM-filtering \cite{vedernikov2025vlm} & VLM-assisted filtering & 68.16 \\
    TCCT-Net \cite{Vedernikov_2024_CVPR} & Hybrid temporal & 68.91 \\
    Transformer \cite{singh2023do} & Transformer-based & 69.10 \\
    ST-GCN \cite{abedi2024eng} & Graph-based & 69.37 \\
    Ordinal ST-GCN \cite{abedi2024eng} & Ordinal graph & 71.24 \\
    \textbf{PriorNet (Ours)} & \textbf{Prior-guided} & \textbf{73.58} \\
    \bottomrule
  \end{tabular}
  }
  \vspace{-0.1in}
\end{table}

PriorNet achieves 73.58\% accuracy on this benchmark, improving over the strongest directly comparable prior result, Ordinal ST-GCN, by 2.34 percentage points. Within the listed public validation-split reports, this is the strongest result. Because the comparison is performed on the public validation partition, these numbers should be interpreted as public-benchmark results rather than hidden-test performance.

\subsection{DAiSEE}

Table~\ref{tab:daisee_method_principle_accuracy} summarizes representative published results on DAiSEE for engagement classification. Because the DAiSEE literature is not fully protocol-uniform, the table should be interpreted as benchmark context rather than as a strictly homogeneous leaderboard. Whenever protocol differences are known or suspected (for example, differences in label handling, modality usage, split construction, or reporting transparency) we do not treat those results as strictly rank-comparable to our setting \cite{malekshahi2024general,shiri2024efficientnetv2,su2024leveraging,singh2024visiophysioenet,gothwal2025vibed,zhu2026vagct}.

\begin{table}[h]
  \centering
  \caption{Representative published results and PriorNet accuracy on DAiSEE for engagement classification. Because reported DAiSEE settings are not fully protocol-uniform across studies, the table should be interpreted as benchmark context rather than as a strictly homogeneous leaderboard.}
  \label{tab:daisee_method_principle_accuracy}
  {\fontsize{8.1}{9.5}\selectfont
  \setlength\tabcolsep{3.5pt}
  \begin{tabular}{@{}l l c@{}}
  \toprule
    \textbf{Method} & \textbf{Principle} & \textbf{Accuracy [\%]} \\
    \midrule
    3D CNN \cite{gupta2016daisee} & End-to-end & 48.60 \\
    Inflated 3D-CNN \cite{zhang2019novel} & End-to-end & 52.40 \\
    HRV + Random Forest \cite{vedernikov2024analyzing} & Remote physiological features & 54.49 \\
    3D-CNN + LSTM \cite{abedi2021improving} & Hybrid temporal & 56.60 \\
    Long-term Recurrent CNN \cite{gupta2016daisee} & End-to-end & 57.90 \\
    Hybrid R(2+1)D and spatio-temporal block \cite{sathisha2024r21d} & Hybrid temporal & 58.62 \\
    ResNet-50 with LSTM with Attention \cite{Liao20216609} & Attention-based & 58.84 \\
    3D-CNN + TCN \cite{abedi2021improving} & Hybrid temporal & 59.97 \\
    Behavioral features + LSTM with attention \cite{huang2019finegrained} & Sequential features & 60.00 \\
    2D ResNet + LSTM \cite{abedi2021improving} & Hybrid temporal & 61.15 \\    
    RefEIP / ModernTCN \cite{LI2024369} & Temporal features & 61.20 \\
    Behavioral features + Neural Turing Machine \cite{ma2021automatic_student_engagement} & Behavioral features & 61.30 \\
    3D DenseNet with Attention \cite{mehta2022densenet_engagement} & Attention-based & 63.59 \\
    ResNet + TCN \cite{abedi2021improving} & Hybrid temporal & 63.90 \\
    ShuffleNet \cite{hu2022optimized_cnn_engagement} & End-to-end & 63.90 \\
    PANet + STformer \cite{su2024leveraging} & Attention-based & 64.00 \\
    EfficientNet + TCN \cite{selim2022students} & Hybrid temporal & 64.67 \\
    Self-Supervised FMAE \cite{zhang2024selfsupervised_engagement} & Self-supervised & 64.74 \\
    EfficientNet + Bi-LSTM \cite{selim2022students} & Hybrid temporal & 66.39 \\
    Supervised Contrastive Ordinal TCN \cite{safa2025superv} & Contrastive ordinal features & 67.32 \\
    Affective/behavioral features + Ordinal TCN \cite{abedi2023affect} & Ordinal features & 67.40 \\
    EfficientNet + LSTM \cite{selim2022students} & Hybrid temporal & 67.48 \\
    \textbf{PriorNet (Ours)} & \textbf{Prior-guided} & \textbf{69.06} \\
    \bottomrule
  \end{tabular}
  }
  \vspace{-0.1in}
\end{table}

PriorNet reaches 69.06\% accuracy on DAiSEE. This places it above all listed prior results in Table~\ref{tab:daisee_method_principle_accuracy}. However, because the DAiSEE literature is not fully protocol-uniform, this comparison should be interpreted as strong benchmark-context evidence rather than as a strictly controlled apples-to-apples ranking across all published studies.

\subsection{DREAMS}

Table~\ref{tab:dreams_method_principle_f1} reports DREAMS results. Because the original benchmark distinguishes single-task, transfer-learning, and multi-task settings, and later work does not always reuse the dataset under a single uniform protocol, the table should be read as protocol-labeled benchmark context rather than as a fully homogeneous leaderboard.

\begin{table}[h]
  \centering
  \caption{Comparative analysis of representative methods and PriorNet weighted F1 on DREAMS.}
  \label{tab:dreams_method_principle_f1}
  {\fontsize{8.1}{9.5}\selectfont
  \setlength\tabcolsep{6pt}
  \begin{tabular}{@{}l l c@{}}
  \toprule
    \textbf{Method} & \textbf{Principle} & \textbf{Weighted F1} \\
    \midrule
    Single-task baseline \cite{Singh2024dreams} & Single-task features & 0.305 \\
    Multi-task baseline \cite{Singh2024dreams} & Multi-task features & 0.306 \\
    Transfer learning baseline \cite{Singh2024dreams} & Transfer learning & 0.293 \\
    TCCT-Net \cite{Vedernikov_2024_CVPR} & Hybrid temporal & 0.357 \\
    TCCT-Net + VLM \cite{vedernikov2025vlm} & VLM-assisted filtering & 0.434 \\
    \textbf{PriorNet (Ours)} & \textbf{Prior-guided} & \textbf{0.515} \\
    \bottomrule
  \end{tabular}
  }
  \vspace{-0.1in}
\end{table}

Within this benchmark framing, PriorNet attains a weighted F1 of 0.515, compared with 0.434 for the strongest listed prior result. The margin is substantial and supports the usefulness of the proposed design on a small, naturalistic dataset with self-reported labels. This pattern is consistent with the design priorities of PriorNet. On a small naturalistic benchmark with subjective self-reports, explicit handling of weak facial evidence and uncertainty-aware optimization are plausible contributors to robustness, although the component-level contribution of each module is analyzed separately only in Sec.~\ref{sec:ablation_analysis}.

\subsection{PAFE}
Table~\ref{tab:pafe_method_principle_f1} summarizes PAFE results under participant-disjoint five-fold evaluation. Published PAFE baselines depend strongly on temporal window length, and the original benchmark reports separate 5\,s, 10\,s, and 20\,s settings. Accordingly, the table is best interpreted as benchmark context under the native fold protocol rather than as a fully controlled same-window ranking. We also note that recent PAFE-related studies \cite{jimaging10050097,buhler2025labwild} report results under different protocols/metrics and are therefore not directly comparable.

\begin{table}[h]
  \centering
  \caption{Comparative analysis of representative methods and PriorNet weighted F1 on PAFE.}
  \label{tab:pafe_method_principle_f1}
  {\fontsize{8.1}{9.5}\selectfont
  \setlength\tabcolsep{6pt}
  \begin{tabular}{@{}l l c c@{}}
  \toprule
    \textbf{Method} & \textbf{Principle} & \textbf{Window} & \textbf{Weighted F1} \\
    \midrule
    SVM \cite{Taeckyung2022PAFE} & Classical features & 5 sec & 0.570 \\
    SVM \cite{Taeckyung2022PAFE} & Classical features & 10 sec & 0.610 \\
    SVM \cite{Taeckyung2022PAFE} & Classical features & 20 sec & 0.600 \\
    XGBoost \cite{Taeckyung2022PAFE} & Boosted features & 5 sec & 0.620 \\
    XGBoost \cite{Taeckyung2022PAFE} & Boosted features & 10 sec & 0.640 \\
    XGBoost \cite{Taeckyung2022PAFE} & Boosted features & 20 sec & 0.650 \\
    DNN \cite{Taeckyung2022PAFE} & Neural features & 5 sec & 0.620 \\
    DNN \cite{Taeckyung2022PAFE} & Neural features & 10 sec & 0.640 \\
    DNN \cite{Taeckyung2022PAFE} & Neural features & 20 sec & 0.680 \\
    TCCT-Net \cite{Vedernikov_2024_CVPR} & Hybrid temporal & -- & 0.720 \\
    TCCT-Net + VLM \cite{vedernikov2025vlm} & VLM-assisted filtering & -- & 0.740 \\
    \textbf{PriorNet (Ours)} & \textbf{Prior-guided} & \textbf{--} & \textbf{0.780} \\
    \bottomrule
  \end{tabular}
  }
  \vspace{-0.1in}
\end{table}

PriorNet reaches a weighted F1 of 0.780, exceeding the strongest listed prior result by 0.040. This is encouraging evidence that the method remains effective under participant-disjoint cross-validation. This result is also encouraging from a generalization perspective, since the evaluation is participant-disjoint and therefore less sensitive to subject-specific memorization. In this setting, the gain is consistent with the broader claim that PriorNet improves robustness rather than optimizing to a single benchmark formulation.

\subsection{Cross-Dataset Summary}
Table~\ref{tab:cross_dataset_summary} provides a protocol-aware summary of the strongest listed prior result and PriorNet on each benchmark. Across the four benchmarks, PriorNet consistently improves over the strongest previously reported comparable results under each dataset's native evaluation protocol. On EngageNet and DAiSEE, where classification accuracy is the primary metric, PriorNet yields absolute gains of 2.34 and 1.58 percentage points, respectively, over the strongest prior baselines. On DREAMS and PAFE, where weighted F1 is the more informative metric due to protocol design and class imbalance, PriorNet improves the best previously reported results by 0.081 and 0.040, respectively.

\begin{table}[h]
  \centering
  \caption{Cross-dataset summary of PriorNet against the strongest prior comparable result on each benchmark.}
  \label{tab:cross_dataset_summary}
  {\fontsize{8.0}{9.3}\selectfont
  \setlength\tabcolsep{4pt}
  \begin{tabular}{@{}l l l c c c@{}}
    \toprule
    \textbf{Dataset} & \textbf{Best Prior Principle} & \textbf{Metric} & \textbf{Best Prior} & \textbf{PriorNet} & \textbf{Gain} \\
    \midrule
    EngageNet & Ordinal graph \cite{abedi2024eng} & Accuracy [\%] & 71.24 & \textbf{73.58} & \textbf{+2.34} \\
    DAiSEE & Hybrid temporal  \cite{selim2022students} & Accuracy [\%] & 67.48 & \textbf{69.06} & \textbf{+1.58} \\
    DREAMS & VLM-assisted filtering \cite{vedernikov2025vlm} & Weighted F1 & 0.434 & \textbf{0.515} & \textbf{+0.081} \\
    PAFE & VLM-assisted filtering \cite{vedernikov2025vlm} & Weighted F1 & 0.740 & \textbf{0.780} & \textbf{+0.040} \\
    \bottomrule
  \end{tabular}
  }
  \vspace{-0.1in}
\end{table}

These results indicate that the proposed prior-guided design generalizes across heterogeneous engagement benchmarks rather than overfitting to a single dataset formulation. In particular, the largest gain on DREAMS suggests that the combination of missing-cue encoding and uncertainty-aware optimization is especially beneficial in settings with limited data and noisier self-reported labels. At the same time, because the four datasets differ in label definitions, split strategies, and primary metrics, the purpose of this summary is not to collapse them into a single aggregate score, but to highlight the consistent direction of improvement obtained by PriorNet across diverse evaluation regimes. The component-level analyses in Sec.~\ref{sec:ablation_analysis} further support this reading by showing that the observed gains are not tied to a single isolated module, but arise from a consistent interaction between preprocessing, adaptation, and objective-level priors.


\section{Ablation and Analysis}
\label{sec:ablation_analysis}

Beyond the benchmark results in Sec.~\ref{sec:results}, we provide component-level and diagnostic analysis of PriorNet. The goal of this section is to determine whether the observed performance gains can be attributed to the intended design choices rather than to an undifferentiated pipeline effect. The component ablation study in Sec.~\ref{sec:component_ablation} is reported on EngageNet and DAiSEE under the same evaluation settings used in the corresponding main benchmark comparisons, while the diagnostic placeholder analysis in Sec.~\ref{sec:placeholders_analysis} is reported on EngageNet only.

We focus on the three design components that define PriorNet: (i) zero-frame placeholders in preprocessing, (ii) a prior-guided low-rank adaptation module (Prior-LoRA) for parameter-efficient backbone adaptation, and (iii) the combined uncertainty-aware training objective based on the Dirichlet-evidential, uncertainty-weighted, and auxiliary cross-entropy terms. The following subsections first isolate their contributions through component ablation across two benchmark settings and then provide a targeted EngageNet analysis of the missing-face regime that the placeholder mechanism is designed to address.

\subsection{Component Ablation on EngageNet and DAiSEE}
\label{sec:component_ablation}

To isolate the contribution of each design choice, we perform a component ablation study on EngageNet and DAiSEE under the same evaluation settings as the corresponding main benchmark comparisons. We define the baseline as a frozen SVFAP backbone with standard face-crop preprocessing, a lightweight classifier head, and conventional cross-entropy training, i.e., without zero-frame placeholders, without Prior-LoRA, and without the advanced objective. Starting from this baseline, we enable the three PriorNet components individually and in combination.

\begin{table}[h]
  \centering
  \caption{Component ablation of PriorNet on EngageNet and DAiSEE. 
  $\checkmark$ indicates that the component is enabled. 
  The baseline uses standard preprocessing, frozen SVFAP features with a classifier head, and cross-entropy training. Results are reported as Accuracy [\%] in the format EngageNet / DAiSEE.}
  \label{tab:component_ablation_engagenet_daisee}
  {\fontsize{8.0}{9.3}\selectfont
  \setlength{\tabcolsep}{3.5pt}
  \begin{tabular}{@{}l c c c c@{}}
    \toprule
    \textbf{Variant} & \textbf{Placeholders} & \textbf{Prior-LoRA} & \makecell{\textbf{Advanced}\\\textbf{objective}} & \makecell{\textbf{Accuracy [\%]}\\\textbf{(EngageNet /}\\\textbf{DAiSEE)}} \\
    \midrule
    Baseline SVFAP + CE                  & --          & --          & --          & 68.91 / 64.46 \\
    + zero-frame placeholders            & \checkmark  & --          & --          & 71.15 / 66.20 \\
    + Prior-LoRA                        & --          & \checkmark  & --          & 70.03 / 65.53 \\
    + advanced objective                 & --          & --          & \checkmark  & 69.56 / 65.02 \\
    + placeholders + Prior-LoRA         & \checkmark  & \checkmark  & --          & 72.36 / 67.54\\
    + placeholders + advanced objective & \checkmark  & --          & \checkmark  & 71.90 / 67.10 \\
    + Prior-LoRA + advanced objective   & --          & \checkmark  & \checkmark  & 70.87 / 66.03 \\
    \textbf{Full PriorNet}              & \checkmark  & \checkmark  & \checkmark  & \textbf{73.58 / 69.06} \\
    \bottomrule
  \end{tabular}
  }
\end{table}

Table~\ref{tab:component_ablation_engagenet_daisee} shows that all three components contribute positively relative to the common baseline on both datasets, while the full model achieves the strongest overall performance in each case. On EngageNet, the largest single gain is obtained by enabling zero-frame placeholders, which improves accuracy from 68.91\% to 71.15\%. This supports the central hypothesis that missing-face events should not be discarded as preprocessing failures, but preserved as potentially informative evidence.

Prior-LoRA also improves over the EngageNet baseline, increasing accuracy to 70.03\%. Although this gain is smaller than that of placeholders, it indicates that parameter-efficient adaptation inside the backbone is preferable to relying on frozen features and a classifier head alone. The advanced objective yields a further positive effect, reaching 69.56\%, which is consistent with the view that uncertainty-aware training can provide useful regularization even under hard-label supervision.

The same qualitative pattern is observed on DAiSEE. Here again, zero-frame placeholders provide the largest single-component gain, increasing accuracy from 64.46\% to 66.20\%. Prior-LoRA and the advanced objective also improve over the common baseline, reaching 65.53\% and 65.02\%, respectively. As on EngageNet, these results indicate that the three components are individually beneficial, with explicit missing-face encoding providing the strongest single contribution.

The pairwise combinations further suggest that the three components are complementary rather than redundant. In particular, combining placeholders with Prior-LoRA yields 72.36\% on EngageNet and 67.54\% on DAiSEE, while combining placeholders with the advanced objective yields 71.90\% and 67.10\%, respectively. The combination of Prior-LoRA and the advanced objective without placeholders also improves over the baseline on both datasets, but remains clearly below the variants that include placeholder encoding. This pattern reinforces the observation that explicit treatment of missing-face events is the strongest individual contributor in the reported settings. Viewed together, the two datasets show the same qualitative ordering of components: placeholder encoding contributes the largest single gain, while Prior-LoRA and the advanced objective provide smaller but consistent improvements that become most effective when combined. This cross-dataset consistency is the main takeaway of the ablation study.

The full PriorNet model reaches 73.58\% on EngageNet and 69.06\% on DAiSEE, corresponding to overall gains of 4.67 and 4.60 percentage points over the respective baselines. Taken together, these results support the claim that the performance improvement is not attributable to a single isolated trick; rather, PriorNet benefits from the interaction of preprocessing priors, parameter-efficient adaptation, and uncertainty-aware optimization across two benchmark settings.

\subsection{Placeholders Analysis}
\label{sec:placeholders_analysis}

To better understand when zero-frame placeholders are beneficial, we analyze their effect as a function of the frequency of face-detection failures in the input clips. Our motivation is that missing-face events are not purely technical artifacts in engagement video: they often co-occur with gaze aversion, strong head turns, partial occlusion, or other visually weak states that may still be behaviorally informative. Instead of discarding such frames, PriorNet preserves them as explicit zero-frame signals, allowing missing visual evidence to remain part of the input sequence.

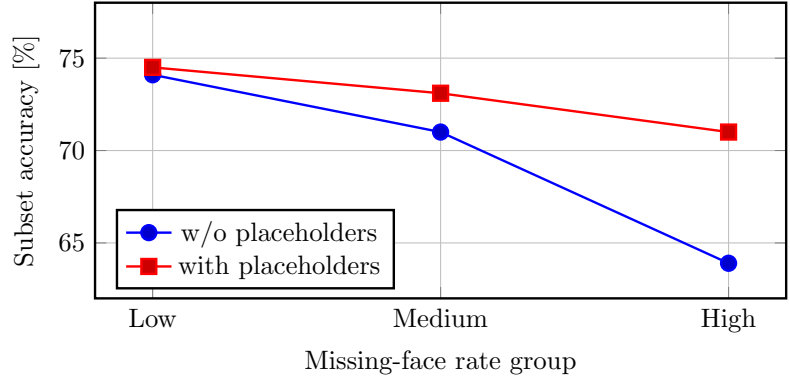
\begin{figure}[h]
\centering
\begin{tikzpicture}
\begin{axis}[
    width=0.82\linewidth,
    height=0.42\linewidth,
    xlabel={Missing-face rate group},
    ylabel={Subset accuracy [\%]},
    symbolic x coords={Low, Medium, High},
    xtick=data,
    ymin=62, ymax=78,
    legend pos=south west,
    grid=major,
    line width=0.9pt,
    mark size=2.8pt
]
\addplot coordinates {(Low,74.1) (Medium,71.0) (High,63.9)};
\addlegendentry{w/o placeholders}

\addplot coordinates {(Low,74.5) (Medium,73.1) (High,71.0)};
\addlegendentry{with placeholders}
\end{axis}
\end{tikzpicture}
\caption{Effect of zero-frame placeholders across clip groups with different missing-face rates on EngageNet. The y-axis reports classification accuracy computed separately within each missingness group. Absolute accuracy is highest in the low-missingness group for both variants, but the relative improvement from placeholder encoding increases as face-detection failures become more frequent.}
\label{fig:placeholders_analysis}
\end{figure}

We partition the EngageNet validation clips into three groups according to the proportion of sampled frames in which the face detector fails: low missing-face rate, medium missing-face rate, and high missing-face rate. For each group, we compute classification accuracy separately within that subset of clips. We then compare the variant without placeholders (\emph{Prior-LoRA + advanced objective}) against the full PriorNet model under otherwise identical settings, so that the difference between the two variants reflects the effect of placeholder encoding.

The results show that both variants perform best in the low-missingness group and degrade as the clips become more challenging. However, the relative benefit of placeholder encoding increases with the missing-face rate. In the low-missingness subset, the gain is small (+0.4 percentage points), indicating that placeholder encoding is less critical when facial evidence is consistently available. In the medium-missingness subset, the gain increases to +2.1 points. The largest improvement is observed in the high-missingness subset, where accuracy increases from 63.9\% to 71.0\% (+7.1 points).

This pattern is consistent with the design rationale of PriorNet: zero-frame placeholders are most useful precisely in the regime where conventional face-centered preprocessing loses the most information. At the same time, this analysis should be interpreted as subset-level diagnostic evidence rather than as a separate benchmark, since the missingness groups differ in difficulty and are not intended as independent evaluation protocols. Together with the two-dataset ablation results, this diagnostic strengthens the interpretation that placeholder encoding is not merely an implementation detail, but a practically important source of robustness in the proposed framework.

\section{Discussion and Limitations}
\label{sec:discussion_limitations}
The results suggest that explicit prior injection is a useful design principle for engagement estimation from face video under the benchmark settings considered in this work. Across four datasets, PriorNet consistently improves over the strongest listed prior reference within each dataset's native evaluation framing, while component ablations on EngageNet and DAiSEE support the view that these gains arise from the combined effect of preprocessing, adaptation, and objective-level priors rather than from a single isolated modification. In this sense, the component trends are not confined to a single benchmark setting. At the same time, these findings should be interpreted as evidence of improved robustness on the evaluated benchmarks, not as a claim that engagement estimation is solved in a general sense.

A first limitation concerns the placeholder mechanism. Zero-frame placeholders depend on the behavior of the face detector, so their benefit may reflect both behaviorally meaningful missing-face events and dataset-specific detector failure patterns. Thus, the placeholder analysis on EngageNet supports the usefulness of explicit missing-face encoding, but does not by itself establish a purely causal interpretation of why the gain arises.

A second limitation is benchmark heterogeneity. The four datasets differ in label definitions, split conventions, primary metrics, and overall protocol structure. Accordingly, the cross-dataset summary should be interpreted as consistent directional evidence across diverse evaluation settings rather than as a single homogeneous leaderboard. 

A third limitation concerns the uncertainty-aware objective. In the current implementation, training is performed with hard labels rather than full annotator-vote distributions. The evidential and uncertainty-weighted terms should therefore be interpreted as regularization and reweighting under hard-label supervision, rather than as direct modeling of annotator disagreement.

Overall, we view PriorNet as a practically useful and methodologically coherent framework for improving facial-video engagement estimation under missing visual evidence, limited adaptation budget, and subjective supervision, while recognizing that its claims remain bounded by detector dependence, protocol heterogeneity, and the current form of the supervision signal.

\section{Conclusion}

This work presented PriorNet, a prior-guided framework for engagement estimation from face video that injects task-relevant priors at three stages of the pipeline: preprocessing, model adaptation, and objective design. Concretely, PriorNet encodes face-detection failures as zero-frame placeholders, adapts a frozen SVFAP backbone through a prior-guided low-rank adaptation module (Prior-LoRA), and trains with a Dirichlet-evidential, uncertainty-weighted objective under hard-label supervision.

Across EngageNet, DAiSEE, DREAMS, and PAFE, PriorNet improved over the strongest listed prior reference within each dataset's evaluation framing. In addition, the component ablations on EngageNet and DAiSEE, together with the EngageNet placeholder analysis, support the view that the gains are not due to a single isolated modification: across the reported ablation settings, explicit missing-face encoding provides the strongest individual contribution, while further improvements arise from combining it with parameter-efficient backbone adaptation and uncertainty-aware optimization.

Taken together, these findings support the central claim of the paper: under the evaluated benchmark conditions, engagement estimation from face video benefits from explicit prior injection rather than relying on backbone scaling alone. At the same time, the scope of this conclusion remains bounded by detector-dependent placeholder construction, hard-label supervision, and the protocol heterogeneity of existing engagement benchmarks. Within those limits, PriorNet provides a practically useful and methodologically coherent framework for improving facial-video engagement estimation under missing visual evidence, limited adaptation budget, and subjective annotation regimes.

\section*{Declarations}
\noindent \textbf{Funding}: This research received no external funding. \vspace{.1in}

\noindent \textbf{Author contributions}: Alexander Vedernikov: Conceptualization, Methodology, Investigation, Data Curation, Software, Analysis, Data Interpretation, Visualization, Writing - Original Draft. \vspace{.1in}

\noindent \textbf{Competing Interests}: The author has no competing interests to declare. \vspace{.1in}

\noindent \textbf{Data Availability Statement}: The datasets used in this study were obtained from the original authors upon request and are not publicly available due to data sharing restrictions. The author does not have permission to redistribute these datasets. Data can be requested directly from the respective dataset owners.\vspace{.1in}

\noindent \textbf{Ethical Approval}: Not applicable. \vspace{.1in}

\noindent \textbf{Consent to Publish}: Not applicable.\vspace{.1in}

\noindent \textbf{Consent to Participate}: Not applicable.\vspace{.1in}
\bibliography{ref}

\end{document}